\documentclass[11pt]{article}
\usepackage{coling2016}
\usepackage{times}
\usepackage{latexsym}
\usepackage{url}
\usepackage{amsmath}
\usepackage{amsfonts}
\usepackage{subfigure}
\usepackage{graphicx}
\usepackage{lipsum}
\usepackage{multirow}

\title{An Empirical Evaluation of various Deep Learning Architectures for Bi-Sequence Classification Tasks}
\author{Anirban Laha \\
  IBM Research India \\
  {\tt anirlaha@in.ibm.com} \\\And
  Vikas Raykar \\
  IBM Research India \\
  {\tt viraykar@in.ibm.com} \\}

\date{}
\begin{document}
\maketitle
\begin{abstract}
Several tasks in argumentation mining and debating, question-answering, and natural language inference involve classifying a sequence in the context of another sequence (referred as bi-sequence classification). For several single sequence classification tasks, the current state-of-the-art approaches are based on recurrent and convolutional neural networks. On the other hand, for bi-sequence classification problems, there is not much understanding as to the best deep learning architecture. In this paper, we attempt to get an understanding of this category of problems by extensive empirical evaluation of 19 different deep learning architectures (specifically on different ways of handling context) for various problems originating in natural language processing like debating, textual entailment and question-answering. Following the empirical evaluation, we offer our insights and conclusions regarding the architectures we have considered. We also establish the first deep learning baselines for three argumentation mining tasks.
\end{abstract}

\section{Introduction}
%
%
\blfootnote{
    %
    %
    %
    %
    %
    
    \hspace{-0.65cm} 
     This work is licensed under a Creative Commons 
     Attribution 4.0 International License.
     License details:
     \url{http://creativecommons.org/licenses/by/4.0/}
}

Argumentation mining is a relatively new challenge in corpus-based discourse analysis that involves automatically identifying argumentative structures within a corpus. Many tasks in argumentation mining~\cite{argumentsurvey2015} and debating technologies~\cite{colingdemo2014} involve categorizing a sequence in the context of another sequence. For example, in \emph{context dependent claim detection}~\cite{cdcd2014}, given a sentence, one task is to identify whether the sentence contains a claim relevant to a particular debatable topic (generally given as a context sentence). Similarly in \emph{context dependent evidence detection}~\cite{evidence2015}, given a sequence (possibly multiple sentences), one task is to detect if the sequence contains an evidence relevant to a particular topic. We refer to such class of problems in computational argumentation as \emph{bi-sequence classification} problems---given two sequences $s$ and $c$ we want to predict the label for the target sequence $s$ in the context of another sequence $c$\footnote{In this paper, we shall ignore the subtle distinction between sentence and sequence and both will mean just a text segment composed of words.}. Apart from the debating tasks, several other natural language inference tasks fall under the same paradigm of having a pair of sequences. For example, \emph{recognizing textual entailment}~\cite{snli2015}, where the task is to predict if the meaning of a sentence can be inferred from the meaning of another sentence. Another class of problems originated from question-answering systems also known as \emph{answer selection}, where given a question, a candidate answer needs to be classified as an answer to the question at hand or not. 

Recently, deep learning approaches have obtained very high performance across many different natural language processing tasks. These models can often be trained in an end-to-end fashion and do not require traditional, task-specific feature engineering. For many single sequence classification tasks, the state-of-the-art approaches are based on recurrent neural networks (RNN variants like Long Short-Term Memory (LSTM)~\cite{lstm97} and Gated Recurrent Unit (GRU)~\cite{cho14}) and convolution neural network based models (CNN)~\cite{Kim14}. Whereas for bi-sequence classification, the context sentence $c$ has to be explicitly taken into account when performing the classification for the target sentence $s$. The context can be incorporated into the RNN and CNN based models in various ways. However there is not much understanding in current literature as to the best way to handle context in these deep learning based models. In this paper, we empirically evaluate(see Section ~\ref{sec:experiments}) the performance of five different ways of handling context in conjunction with target sentence(see Section~\ref{sec:models}) for multiple  bi-sequence classification tasks(see Section~\ref{sec:tasks}) using architectures composed of RNNs and(/or) CNNs. 

In a nutshell, this paper makes the two major novel contributions: 
\begin{enumerate}
\item We establish the first deep learning based baselines for three bi-sequence classification tasks relevant to argumentation mining with zero feature engineering.
\item We empirically compare the performance of several ways handling context for bi-sequence classification problems in RNN and CNN based models. While some of these variants are used in various other tasks, there has been no formal comparison of different variants and this is the first attempt to actually list all the variants and compare them on several publicly available benchmark datasets.
\end{enumerate}


\section{Bi-Sequence classification tasks}\label{sec:tasks}
In this section, we will briefly mention the various bi-sequence tasks of interest in the literature of argumentation mining and in the broader natural language inference domain.

\subsection{Argumentation Mining}
We mainly consider two prominent tasks in argumentation mining, namely,  detecting the claims~\cite{cdcd2014} and evidences~\cite{evidence2015}, within a given corpus, which are related to a pre–specified topic. These two tasks together helps to automatically construct persuasive arguments out of a given corpora. We will define the following four concepts:

\noindent \textbf{Motion} -  The topic under debate, typically a short phrase that frames the discussion.

\noindent \textbf{Claim} - A general, typically concise statement that directly supports or contests the motion.

\noindent\textbf{Motion text} - A document/article/discourse that contain claims with high probability.

\noindent \textbf{Evidence} - A set of statements that directly supports the claim for a given motion.

\subsubsection{Context Dependent Claim Detection (CDCD)}
\label{sec:cdcd}
\emph{Given a sentence in a motion text the task is to identify whether the sentence contains a claim relevant to the motion or not}. This is the claim sentence task introduced by ~\newcite{cdcd2014}. For example, each of the following sentences includes a claim, marked in italic, for the motion topic in brackets.
\begin{enumerate}
  \item \small (\textbf{the sale of violent video games to minors}) Recent research has suggested that some  \emph{violent video games may actually have a pro-social effect in some contexts, for example, team play}.
  \item (\textbf{the right to bear arms}) Some gun control organizations say that \emph{increased gun ownership leads to higher levels of crime, suicide and other negative outcomes}.
\end{enumerate}

\subsubsection{Context Dependent Evidence Detection (CDED)}
\label{sec:cded}
\emph{Given a segment in a motion text the task is to identify whether the segment contains an evidence relevant to the motion or not}~\cite{evidence2015}. We consider evidences of two types in this paper, \emph{Study} and \emph{Expert}. Evidences of type study are generally results of a quantitative analysis of data given as numbers, or as conclusions. The following are two examples for study evidence relevant to the motion topic in brackets.
\begin{itemize}
  \item \small (\textbf{the sale of violent video games to minors}) A 2001 study found that exposure to violent video games causes at least a temporary increase in aggression and that this exposure correlates with aggression in the real world.
  \item (\textbf{the right to bear arms}) In the South region where there is the highest number of legal guns per citizen only $59\%$ of all murders were caused by firearms in contrast to $70\%$ in the Northeast where there is the lowest number of legal firearms per citizen.
\end{itemize}
Evidence of type expert is a testimony by a person/group/commitee/organization with some known expertise/authority on the topic. The following are two examples for expert evidence relevant to the motion topic in brackets.
\begin{enumerate}
  \item \small (\textbf{the sale of violent video games to minors}) This was also the conclusion of a meta-analysis by psychologist Jonathan Freedman, who reviewed over 200 published studies and found that the majority did not find a causal link.
  \item (\textbf{the right to bear arms}) University of Chicago economist Steven Levitt argues that available data indicate that neither stricter gun control laws nor more liberal concealed carry laws have had any significant effect on the decline in crime in the 1990s.
\end{enumerate}

\subsection{Textual Entailment (TE)}
This task \cite{snli2015} corresponds to a multiclass setting, where given a pair of sentences (\emph{premise} and \emph{hypothesis}), the task is to identify whether one of them (\emph{premise}) entails, contradicts or is neutral with respect to the other sentence (\emph{hypothesis}). Unlike the other debating tasks seen previously, we cannot call these pair of sentences as context and target as these are more symmetric in nature. Typical examples\footnote{\url{http://nlp.stanford.edu/projects/snli/}} are the following (premise followed by hypothesis):
\begin{itemize}
    \item \small Entailment: A soccer game with multiple males playing - Some men are playing a sport.
    \item Contradiction: A black race car starts up in front of a crowd of people - A man is driving down a lonely road.
    \item Neutral: A smiling costumed woman is holding an umbrella - A happy woman in a fairy costume holds an umbrella.
\end{itemize}

\subsection{Answer Selection for Questions}
Question Answering (QA) System is a natural extension to the traditional commercial search engines as it is concerned with fetching answers to natural language queries and returning the information accurately in natural human language. A QA system can be either closed-domain or open-domain, the former being restricted to a particular domain while the latter is not. \emph{Answer sentence selection is a crucial subtask of the open-domain question answering problem, with the goal of extracting answers from a set of pre-selected sentences} \cite{wikiqa}. This is again bi-sequence classification task where the pair of sequences being a question and a candidate answer to be selected.

\section{Deep Learning models for sequence pairs}\label{sec:models}
All the tasks described in the previous section can be formulated as bi-sequence classification problems where we have to predict the label for the given pair of sequences. For simpler single sequence text classification tasks, RNN or CNN based architectures have become standard baselines. In this section, we will briefly introduce RNN and CNN and then subsequently describe RNN and CNN based architectures for bi-sequence classification tasks. Specifically, we talk about five different ways of handling context along with the target sentence.

\subsection{Continuous Bag of Words (CBOW)}\label{sec:cbow}
One of the simplest forms of sequence representation is the CBOW model, where every word in the sequence produces some word embedding (say, based on word2vec \cite{word2vec13}) and the average of the word embedding vectors over the words produces the representation of the sequence. As is evident, this form of representation totally disregards the word order of the sequence.

\subsection{Recurrent neural networks (RNNs)}
The RNN model provides a framework for conditioning on the entire history of the sequence without resorting to the Markov assumption traditionally used for modelling sequences. Unlike CBOW, RNNs encode arbitrary length sequences as fixed size vectors without disregarding the word order. 

Given an ordered list of $n$ input vectors $\textbf{x}_1,...,\textbf{x}_n$ and an initial state vector $\textbf{s}_0$, a RNN generates an ordered list of $n$ state vectors $\textbf{s}_0,...,\textbf{s}_n$ and an ordered list of $n$ output vectors $\textbf{y}_1,...,\textbf{y}_n$, that is, $RNN(\textbf{s}_0,\textbf{x}_1,...,\textbf{x}_n)=\textbf{s}_1,...,\textbf{s}_n,\textbf{y}_1,...,\textbf{y}_n$. The input vectors $\textbf{x}_i$ (which corresponds to a fixed dimensional representation for each word in the sequence) are presented to the RNN in a sequential fashion and $\textbf{s}_i$ represents the state of the RNN after observing the inputs $\textbf{x}_1,...,\textbf{x}_i$. The output vector $\textbf{y}_i$ is a function of the corresponding state vector $\textbf{s}_i$ and is then used for further prediction. An RNN is given by the following update equations:
\begin{eqnarray}
  \textbf{s}_i &=& R(\textbf{x}_i,\textbf{s}_{i-1}) \\
  \textbf{y}_i &=& O(\textbf{s}_{i})
\end{eqnarray}
The recursively defined function $R$ takes as input the previous state vector $\textbf{s}_{i-1} $ and the current input vector $\textbf{x}_i \in \mathbb{R}^{d_{x}}$ and results in an updated state vector $\textbf{s}_i  \in \mathbb{R}^{d_{s}}$. An additional function $O$ maps the state vector $\textbf{s}_i$ to an output vector $\textbf{y}_i \in \mathbb{R}^{d_{y}}$.  Different instantiations of $R$ and $O$ will result in the different network structures (Simple RNN, LSTM~\cite{lstm97}, GRU~\cite{cho14}, etc.).
The final state vector $\textbf{s}_n$ can be thought of as encoding the entire input sequence into a fixed size vector, which can be passed to a softmax layer to produce class probabilities.

\subsection{Convolutional Neural Networks (CNNs)}
CNNs are built on the premise of locality and parameter sharing which has proven to produce very effective feature representation for images. Following the groundbreaking work by \newcite{Kim14}, there has been a lot of interest shown by the text community towards applying CNNs for modelling text representation. 

As in case of RNNs defined above, an $n$-word sentence consists of embedding vectors $\textbf{x}_1,...,\textbf{x}_n \in \mathbb{R}^{d_{x}}$, one for each word in the sentence. Let $\textbf{x}_{i:i+j}$ denote the concatenation of words $\textbf{x}_i,\textbf{x}_{i+1},...,\textbf{x}_{i+j}$. A convolution operation defined by a non-linear function $f$ applies a filter $\textbf{w} \in \mathbb{R}^{h d_{x}}$ to a window of $h$ words to produce a single feature value as given below:
\begin{eqnarray}
    c_i = f (\textbf{w} . \textbf{x}_{i:i+h-1} + b) \\
    \textbf{c} = [c_1,c_2,...,c_{n-h+1}] 
\end{eqnarray}

In the next step, max-pooling is applied which essentially produces a single feature value $\hat{c} = max\{\textbf{c}\}$, corresponding to one filter that has been used. The model can have multiple feature values, one for each applied filter, thus producing a feature representation for the input sentence, which can again be passed to a softmax layer to produce class probabilities.

\subsection{Bi-Sequence RNN models}
\label{sec:biseq-rnn}
\begin{figure*}[t]
\centering
\subfigure[conditional-state]{\includegraphics[width=0.32\textwidth]{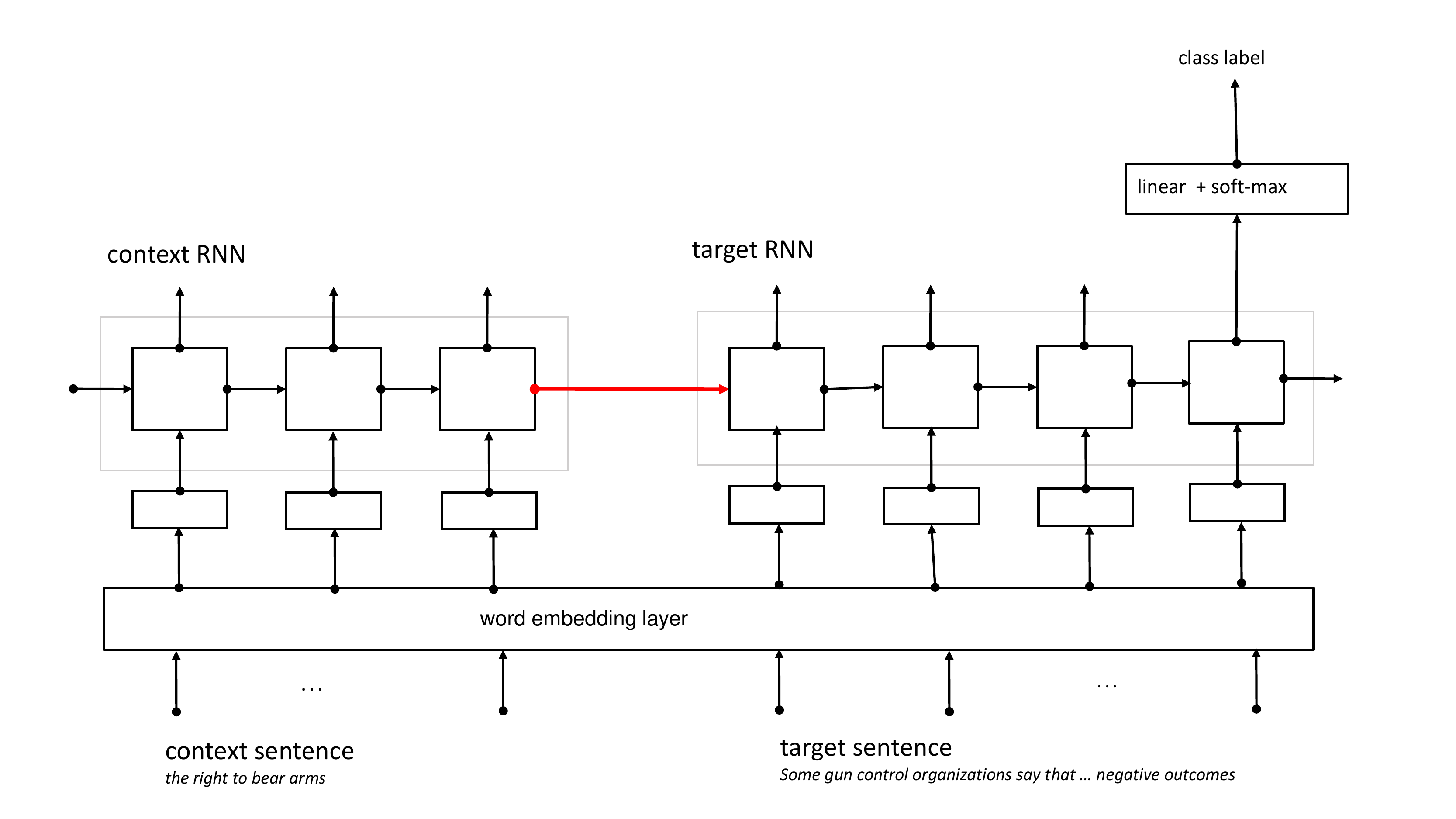}}
\subfigure[conditional-input]{\includegraphics[width=0.32\textwidth]{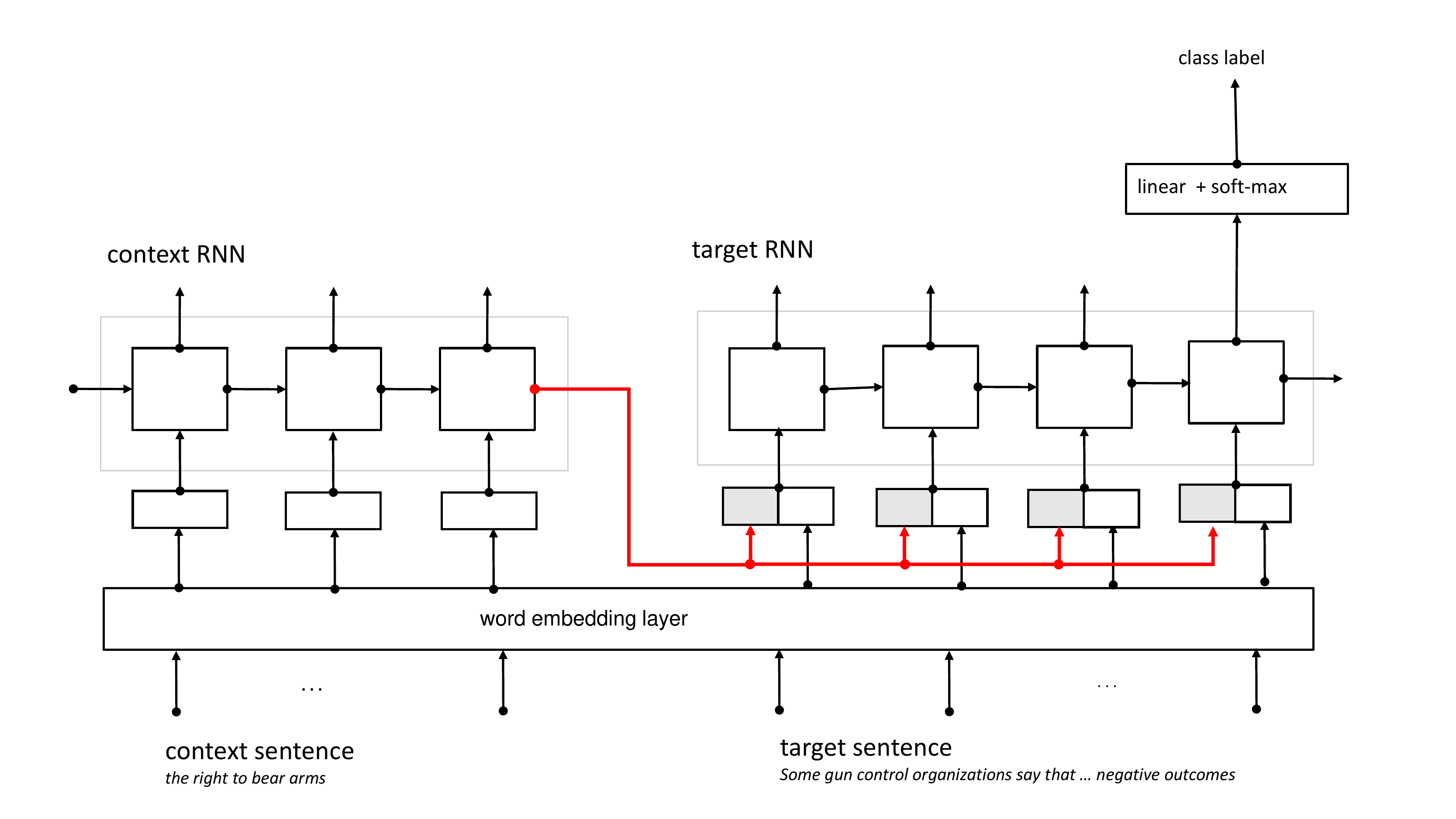}}
\subfigure[conditional-state-input]{\includegraphics[width=0.32\textwidth]{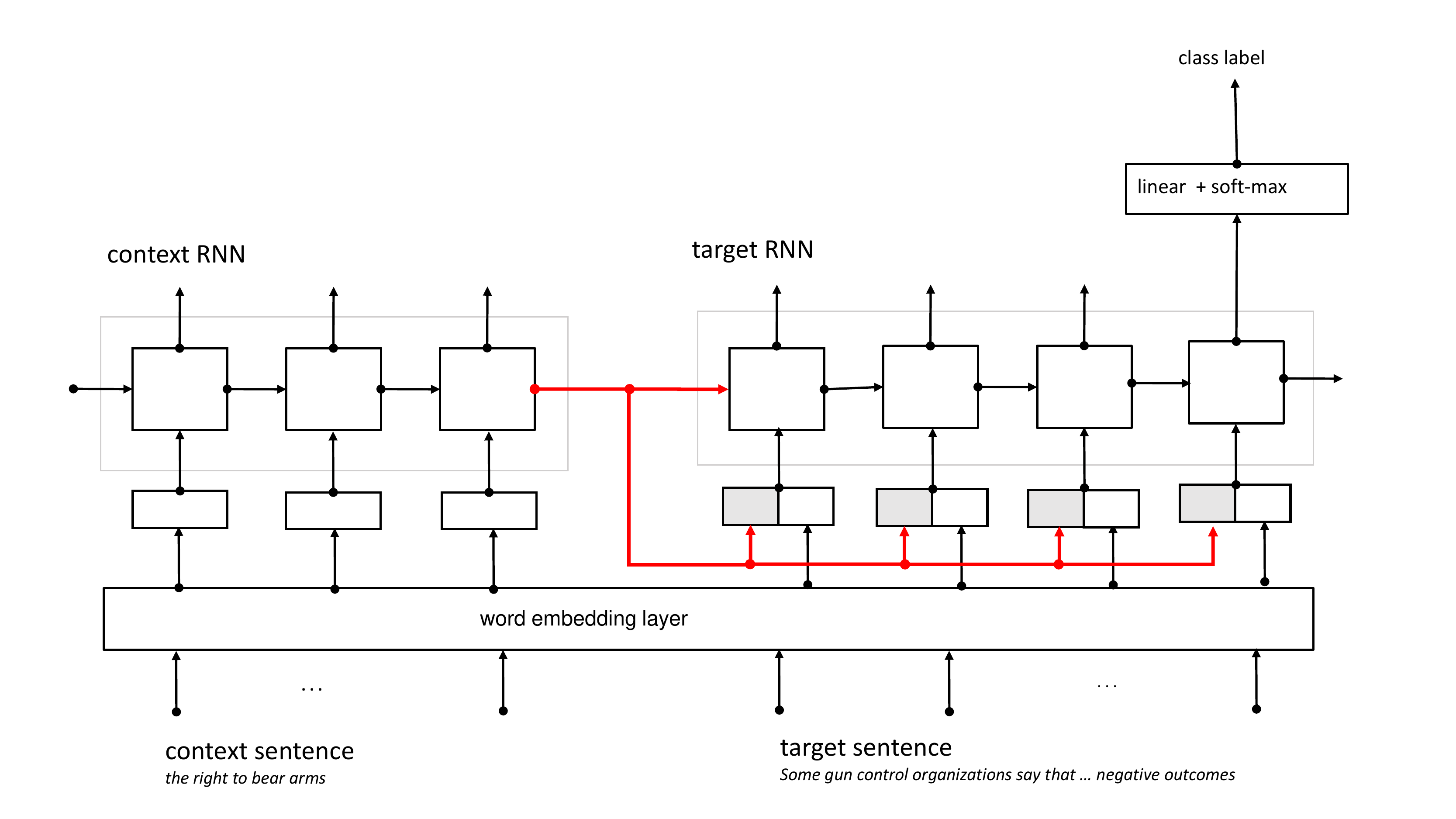}}
\subfigure[concat]{\includegraphics[width=0.32\textwidth]{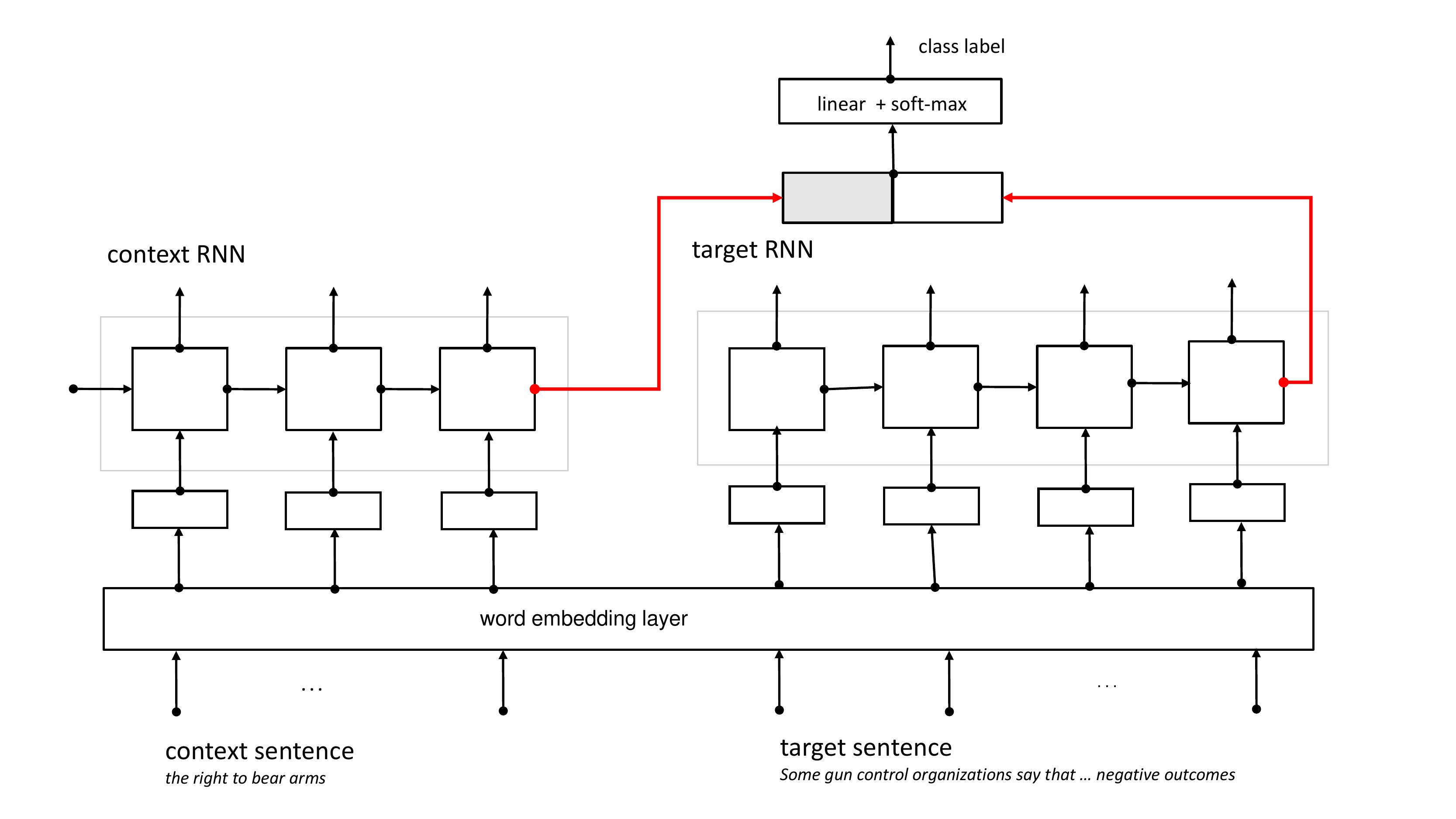}}
\subfigure[bilinear]{\includegraphics[width=0.32\textwidth]{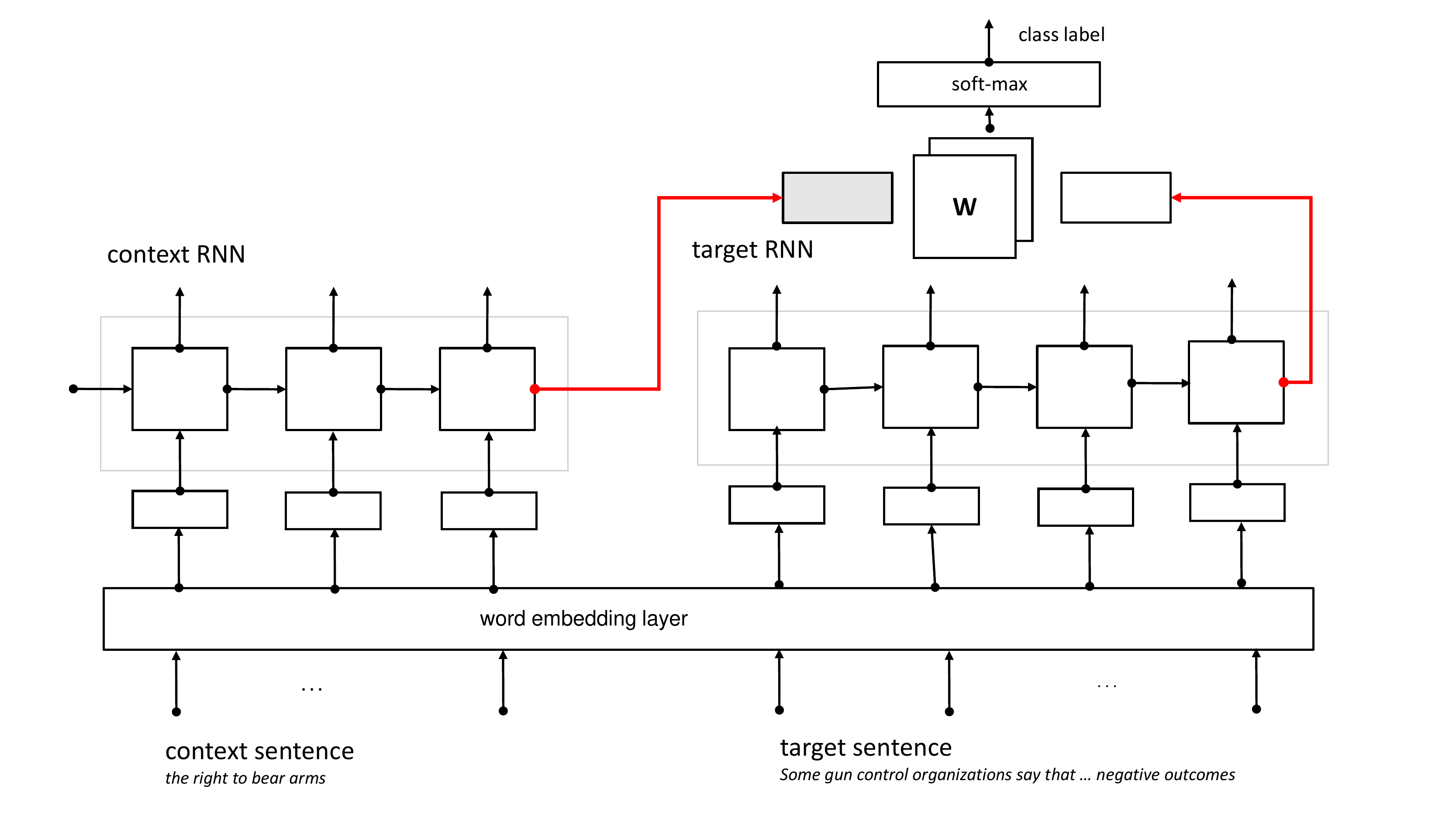}}
\subfigure[concat-sentence]{\includegraphics[width=0.32\textwidth]{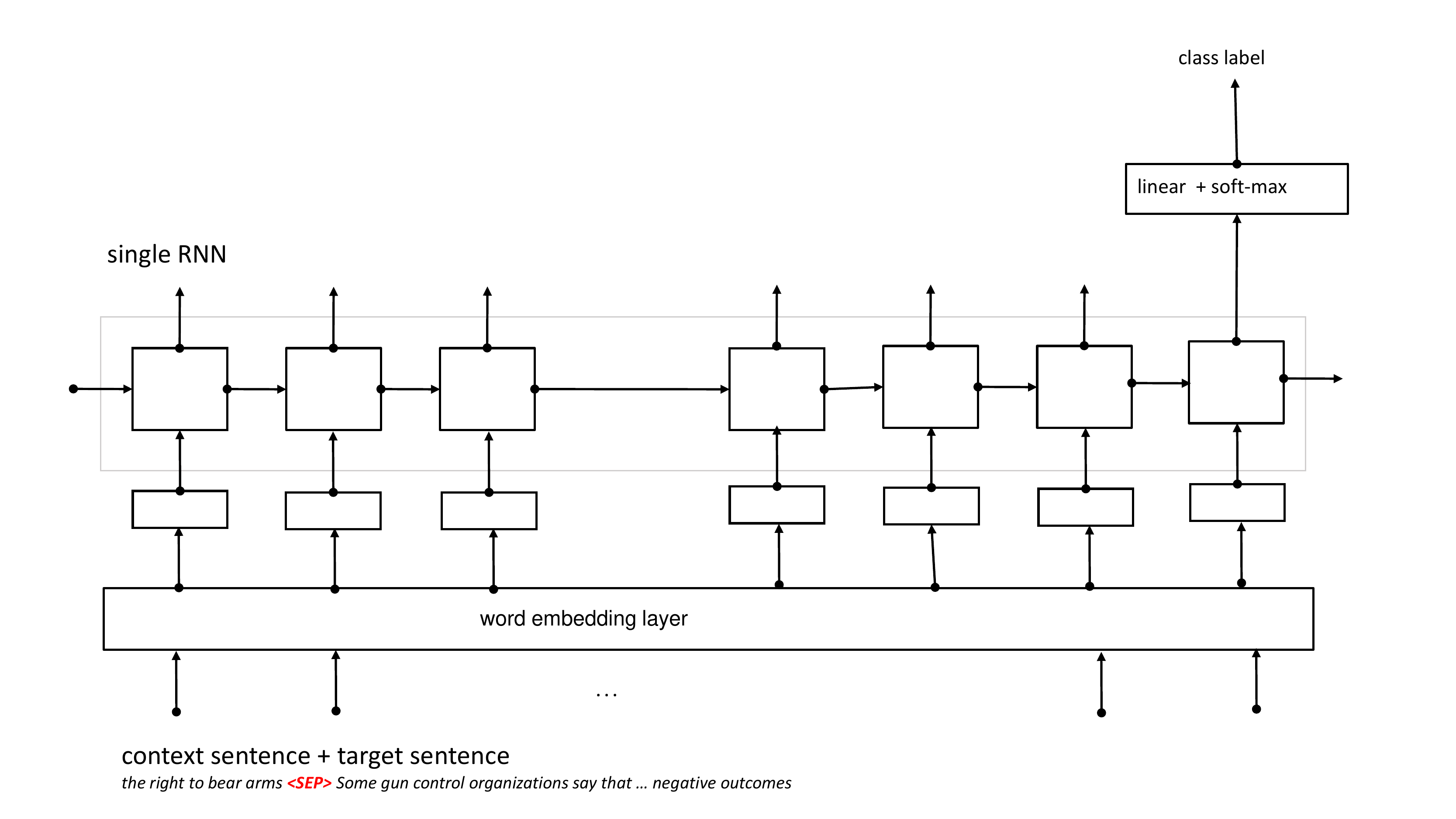}}
\caption{RNN based Architectures for bi-sequence classification.}\label{fig:models}
\end{figure*}

For bi-sequence classification tasks we use two RNNs, one RNN to encode the context sentence (\emph{context RNN}) and another separate RNN encode the target sentence (\emph{target RNN}). We define the following five different variants of combining these two RNNs for bi-sequence classification tasks (see Figure~\ref{fig:models} for illustration of these variants). 
\begin{enumerate}
  \item \small \textbf{conditional-state}: The final state of the context RNN is fed as the initial state of the target RNN. This way of handling context for RNNs has been previously used in conversational systems~\cite{Vinyals:15}, image description~\cite{vinyals2015show} and image question answering~\cite{Mengye:15} systems.
  \item \textbf{conditional-input}: The final state of the context RNN is fed as auxiliary input (concatenated with every input) for the target RNN. This way of handling context has been previously used in machine translation tasks~\cite{sutskever2014sequence}.
  \item \textbf{conditional-state-input}: The final state of the context RNN is fed as the initial state of the target RNN and also fed as input for target RNN concatenated with every input.
  \item \textbf{concat}: The final states of both the context and the target RNN are concatenated and then fed to a softmax layer for the label prediction.
  \item \textbf{bi-linear}: The final states of both the context and the target RNN are combined using a bi-linear form ($\textbf{x}^{\top}\textbf{W}\textbf{y}$) with a softmax function for the final prediction. There are different $\textbf{W}$ for different classes under consideration.
\end{enumerate}
From here on, we would refer to architecture types 1, 2 and 3 as \emph{conditional} variants while the others will be addressed as is.
In addition, we consider another baseline variant \textbf{concat-sentence}, in which we concatenate the context and the sentence with a special separator token and feed the entire concatenated sequence to a single RNN.
For all these variants we use a common embedding layer. Also note that the conditional variants require a common RNN size for both the context and the target RNNs. Even though that restriction is not there for other variants, we choose the same RNN size anyways for convenience.

\begin{figure*}[t]
\centering
\subfigure[concat-bilinear-all-combinations]{\label{fig:models11}\includegraphics[width=0.45\textwidth]{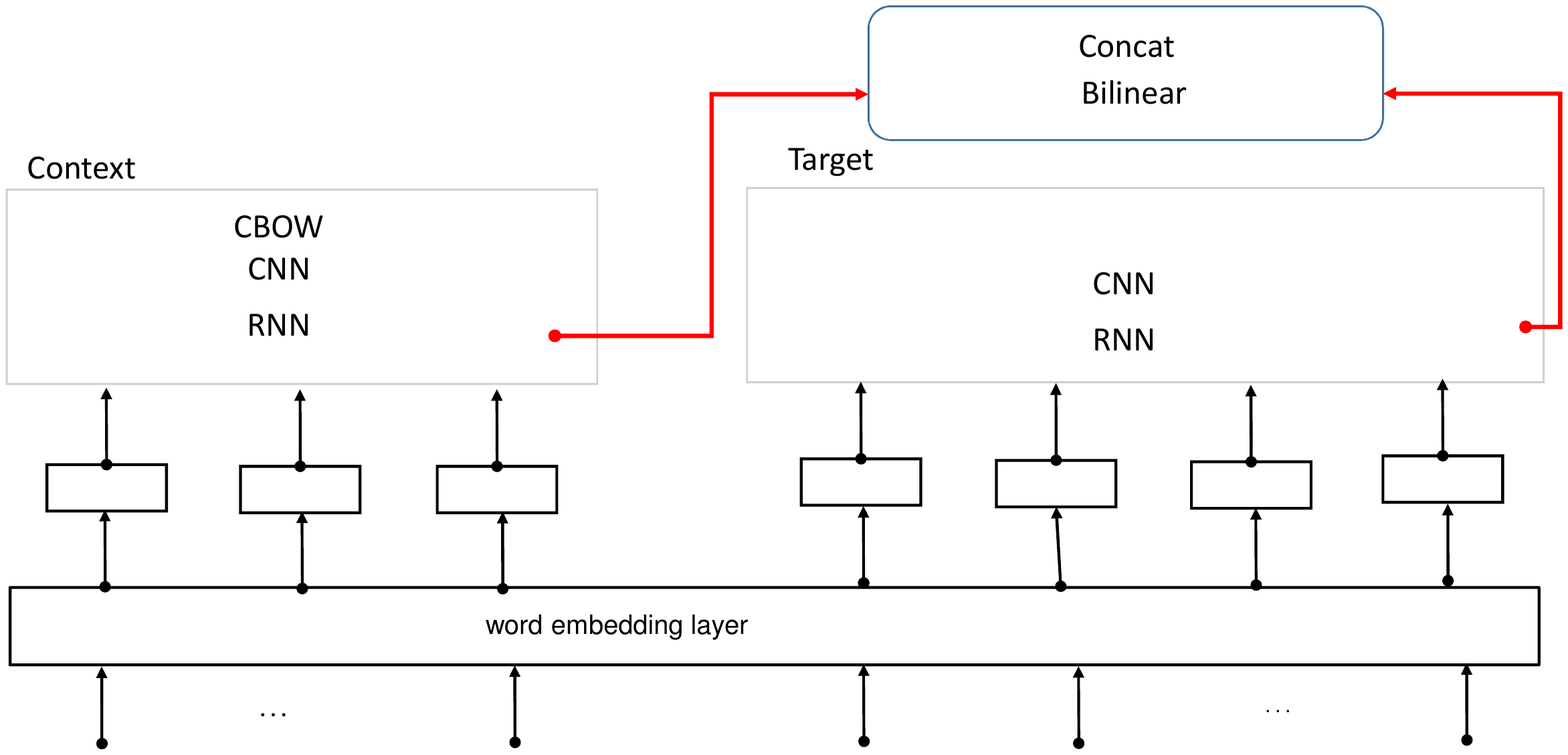}}
\subfigure[conditional-all-combinations]{\label{fig:models12}\includegraphics[width=0.45\textwidth]{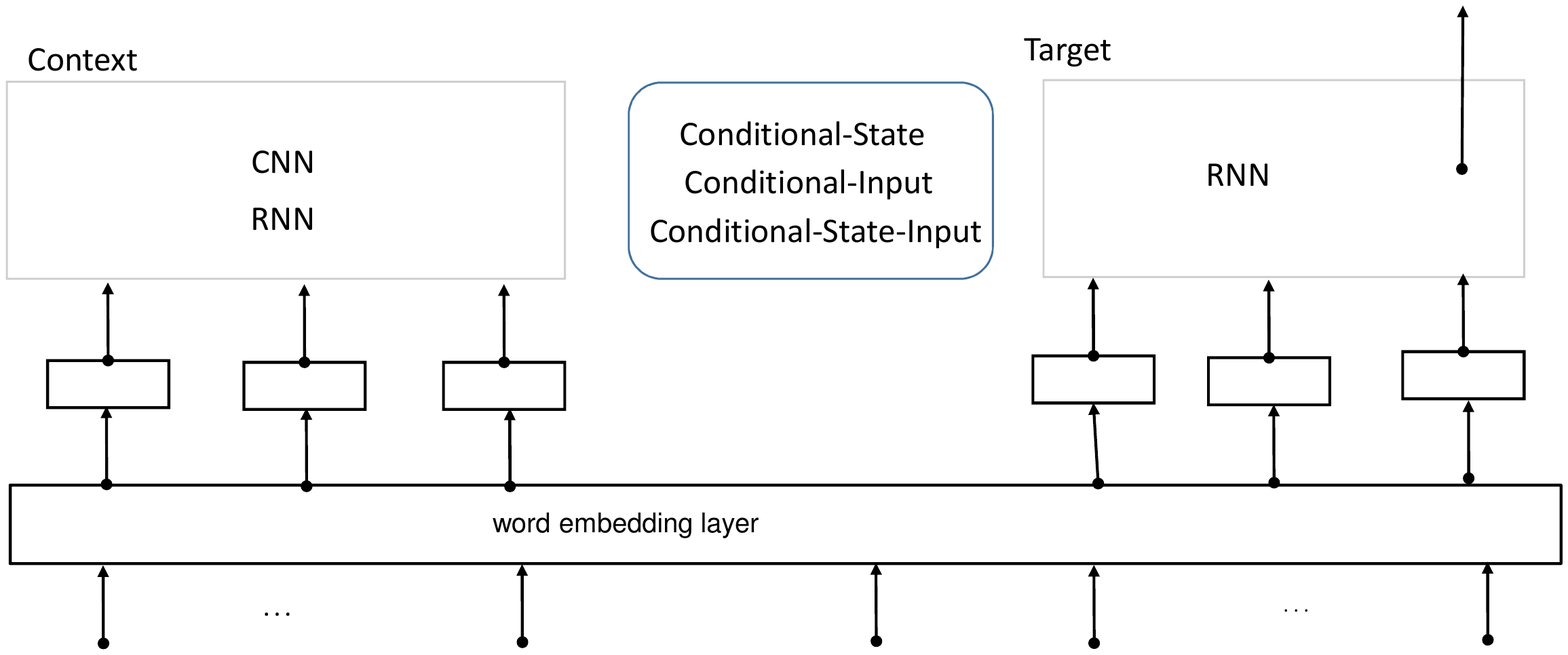}}
\caption{Multiple model variants for  bi-sequence classification.}\label{fig:models1}
\end{figure*}

\subsection{Bi-Sequence model variants}
\label{sec:biseq-all}
In this paper, we consider multiple ways of extending the bi-sequence architectures mentioned in section ~\ref{sec:biseq-rnn}, by replacing RNN with CBOW or CNN either for context or target or both. For the variations \emph{concat} and \emph{bi-linear} (see Fig. ~\ref{fig:models11}), we have considered CBOW, RNN and CNN for context representation whereas we have RNN and CNN for target representation. In tasks where context has very few words (say debating tasks or question-answering), a simple representation like CBOW may work for context. However, we haven't considered modelling target using CBOW as targets are usually of larger length. For the \emph{conditional} variants (see Fig. ~\ref{fig:models12}), we haven't considered CBOW due to their limited modelling capacity and there is no softmax layer directly on top of context representation to compensate for it (even though softmax is only on top of target RNN). Moreover, target can only be RNN as there is no concept of hidden state for CNNs. Hence, we use CNN and RNN for context whereas RNN for target. In addition, we consider the concat-sentence (mentioned in section ~\ref{sec:biseq-rnn}) as a baseline. This leads to 19 architectures for empirical comparison (12 from Fig. ~\ref{fig:models11} and 6 from Fig. ~\ref{fig:models12} and the baseline).


\section{Experiments}
\label{sec:experiments}
We have carried out extensive evaluation of the above architecture variants over a wide range of datasets related to argumentation mining as well as datasets appealing to the larger natural language community like textual entailment and question answering data. We consider data with class imbalance problem as well as balanced and we do not restrict ourselves to binary classification by working with multiclass dataset as well. As can be found in the Table~\ref{debater-data}, we consider the following tasks related to the domain of argumentation mining\cite{aharoni14}, which is available here \footnote{\url{https://www.research.ibm.com/haifa/dept/vst/mlta_data.shtml}} :
\begin{itemize}
\item \small Claim Sentence : This is the dataset for the CDCD task defined in section ~\ref{sec:cdcd}. This is the current benchmark dataset for the Claim Detection task.
There are a total 47183 canditate claims distributed among 33 motions.
\item EXPERT Evidence : This is corresponding to the CDED task defined in section ~\ref{sec:cded} for evidence type EXPERT. There are 56985 labelled candidates for 57 different motion topics.
\item STUDY Evidence : For evidence type STUDY, the dataset consists of 33534 labelled candidates for 49 motion topics. 
\end{itemize}

Table~\ref{debater-data} summarizes all of the datasets above. Interesting point to be noted here is that all the datasets above have very low number of positives and the architectures we are evaluating need to be resilient to the class imbalance problem for these datasets. 
Other than the debating datasets listed above, we also consider two datasets related to more popular problems in the natural language processing community:
\begin{itemize}
\item \small Textual Entailment (TE) \footnote{\url{http://nlp.stanford.edu/projects/snli/}} \cite{snli2015} :This dataset consist of around 500K instances evenly distributed across all three classes. So, here we have a multiclass problem in a balanced setting.
\item{WikiQA \footnote{\url{http://aka.ms/WikiQA}} \cite{wikiqa} :  There are around 29K labelled question/answer pairs at our disposal.}
\end{itemize}

\begin{table}[!h]
\small
\parbox{.4\linewidth}{
\centering
\begin{tabular}{|l|l|l|l|}
\hline \bf \small{Task} & \bf \small{Motions} & \bf \small{Data Size} & \bf \small{Positives} \\ \hline
\small{Claim Sentence} & 33 & 47183 & 2.77\% \\
\small{EXPERT Evidence} & 57 & 56985 & 4.56\% \\
\small{STUDY Evidence} & 49 & 33534 & 3.74\% \\
\hline
\end{tabular}
\caption{\label{debater-data} Argument Mining Datasets.}
}
\hfill
\parbox{.5\linewidth}{
\centering
\begin{tabular}{|l|r|l|l|l|l|}
\hline \bf \small{Task} & \bf \small{Train} & \bf \small{Dev} & \bf \small{Test} & \bf \small{Problem} & \bf \small{Class} \\ \hline
\small{TE} & \small{549367} & \small{9842} & \small{9824} & \small{Multiclass}& \small{Balance} \\
\small{WikiQA} & \small{20360} & \small{2733} & \small{6165} & \small{Binary} & \small{Imbalance} \\
\hline
\end{tabular}
\caption{\label{public-data} More Datasets.}
}
\end{table}

\subsection{Experimental Setup}
For each of the architectures mentioned in section ~\ref{sec:biseq-all}, we choose the best configuration of hyperparameters based on the validation portion of the particular dataset (For Claim and Evidence datasets, we consider a train:valid:test split of 60:10:30 while for the TE and WikiQA datasets we consider their corresponding given split). 

Performance of the best performing configuration for every architecture is reported on the test data using the appropriate metric. As the claim and expert and study Evidences had similar data characteristics (in terms of data size and context and target lengths), we did extensive hyperparam tuning only on the claim dataset and applied the best configurations without further tuning to the expert and study datasets. Exhaustive hyperparam tuning was done on the TE dataset as well because its data characteristics are very different from other datasets.

In addition to reporting the test metrics for argumentation datasets, we carried out Leave-One-Out(leaving one motion out for testing) Mode training and evaluation which is more appropriate for this problem setting as it is crucial that we generalize well to totally unseen motion topics. In this case, we report the macro-average metrics over all motions.

\subsection{Hyperparameter Tuning}
Considering the number of variations of combinations of architectures we have considered, we have a huge hyperparameter space to deal with. Hence, we decided to fix insignificant hyperparameters and focus only on the relevant ones. We have decided to use word2vec \cite{word2vec13} pretrained models for initializing the word embeddings across CBOW, CNN and RNNs and made them trainable specific to task at hand. In addition we have found through minimal tuning that the Adam \cite{adam14} optimizer seems to work best. We have also found that a learning rate of 0.001 works best in most scenarios except when the parameter space in some architectures (for ex, \emph{bi-linear}) is large, in which case lower learning rates of 0.0001 or 0.00001 worked well. Rather than tuning the maximum sequence length for context and target sentence, we tried to fix it by getting reasonable values by plotting histogram of sequence lengths and had a cut-off at around  98-99 percentile. The max lengths turned out to be 14 for context and 60 for target for Claim, Evidence and WikiQA datasets while for textual entailment, they turned out to be 30 in both due to the symmetrical nature between premise and hypothesis. We tuned the following hyperparameters:\\
\small
  \textbf{RNN Model} : GRU or LSTM.\\
  \textbf{RNN Size} : 50,100,200,300,400,500,1000.\\
  \textbf{CNN Filter Sizes} : 3,3+4,3+4+5,2+3+4+5.\\
  \textbf{CNN Number of Filters} : 10,20,40,64,128.\\
  \textbf{L2 Reg coeff for CNN} : 0, 0.01, 0.001, 0.0001.\\
\normalsize
For every architecture type, we carried out the optimization of the relevant hyperparams from the above list over the whole grid. One point to note is that for certain architectures like conditional-state, as output of context is fed in to the hidden state of target RNN, there are some restrictions in the allowable context RNN/CNN hyperparam configurations based on the target RNN settings as the output dimension of the context RNN/CNN needs to match the hidden state dimension of the target RNN.

\subsection{Evaluation Metrics}
For the datasets Claim, Expert, Study Evidence and WikiQA datasets, we have used standard evaluation measures like Average Precision (Area under Precision Recall Curve) and AUC to choose best hyperparam configurations based on validation data as well as report test metrics. For argument mining specific tasks, we have reported other additional metrics like P@200, R@200, F1@200, P@50, R@50 and F1@50 \cite{cdcd2014} in addition to reporting AUC and Average Precision. Please note for leave-one-out mode, the reported metrics are macro-average over all motion topics. For Textual entailment, since it is a more balanced dataset, reporting valid and test accuracies are standard in the literature and we have done the same.

\begin{table*}[!ht]
\scriptsize
\centering
  \begin{tabular}{|l|l|l|l|r|}
    \hline
    \bf Task &\bf Context & \bf Target & \bf Architecture & \bf Test AVGP\\
    \hline
    \multirow{3}{*}{Claim Sentence} & RNN & CNN & \bf Concat & \bf 0.307 \\
    \cline{2-5}
    & CNN & CNN & Concat  & 0.304 \\
    \cline{2-5}
    & \multicolumn{3}{l|}{Concat-Sentence baseline} & 0.17 \\
    \hline \hline
    \multirow{3}{*}{EXPERT Evidence} & RNN & RNN & \bf Conditional-State-Input  & \bf 0.257 \\
    \cline{2-5}
    & CNN & CNN & Concat  & 0.254 \\
    \cline{2-5}
    & \multicolumn{3}{l|}{Concat-Sentence baseline} & 0.225 \\
    \hline \hline
    \multirow{3}{*}{STUDY Evidence} & CNN & CNN & \bf Concat & \bf 0.297 \\
    \cline{2-5}
    & RNN & CNN & Concat & 0.29 \\
    \cline{2-5}
    & \multicolumn{3}{l|}{Concat-Sentence baseline} & 0.236 \\
    \hline \hline
    \multirow{2}{*}{WikiQA} & CBOW & RNN & \bf Concat & \bf 0.187 \\
    \cline{2-5}
    & CNN & RNN & Conditional-State-Input & 0.186 \\
    \hline
  \end{tabular}
  \caption{Empirical evaluation based on Average Precision on assymetric datasets.}
  \label{table-avgp-summary}
\end{table*}

\begin{table*}[!ht]
\scriptsize
\centering
  \begin{tabular}{|l|l|l|l|r|}
    \hline
    \bf Task &\bf Context & \bf Target & \bf Architecture & \bf Test AUC\\
    \hline
    \multirow{3}{*}{Claim Sentence} & CNN & CNN & \bf Concat & \bf 0.873 \\
    \cline{2-5}
    & CNN & RNN & Conditional-State  & 0.873 \\
    \cline{2-5}
    & \multicolumn{3}{l|}{Concat-Sentence baseline} & 0.831 \\
    \hline \hline
    \multirow{3}{*}{EXPERT Evidence} & RNN & RNN & \bf Conditional-State  & \bf 0.832 \\
    \cline{2-5}
    & RNN & RNN & Conditional-State-Input  & 0.823 \\
    \cline{2-5}
    & \multicolumn{3}{l|}{Concat-Sentence baseline} & 0.805 \\
    \hline \hline
    \multirow{3}{*}{STUDY Evidence} & CNN & CNN & \bf Concat & \bf 0.87 \\
    \cline{2-5}
    & CBOW & CNN & Concat & 0.864 \\
    \cline{2-5}
    & \multicolumn{3}{l|}{Concat-Sentence baseline} & 0.844 \\
    \hline \hline
    \multirow{2}{*}{WikiQA} & CNN & CNN & \bf Concat & \bf 0.74 \\
    \cline{2-5}
    & CBOW & RNN & Concat & 0.74 \\
    \hline
  \end{tabular}
  \caption{Empirical evaluation based on AUC on assymetric datasets.}
  \label{table-auc-summary}
\end{table*}

\begin{table*}[!ht]
\centering
\resizebox{\textwidth}{!}{\begin{tabular}{|l|l|r|r|}
    \hline
    \bf Method &\bf Model & \bf TrainAcc(\%) & \bf TestAcc(\%) \\
    \hline
    \cite{snli2015} & Use of features incl unigrams and bigrams & 99.7 & 78.2\\
    \cite{vendrov15} & 1024D GRU encoders w/ unsupervised 'skip-thoughts' pre-training & 98.8 & 81.4\\
    \cite{mou15} & 300D Tree-based CNN encoders & 83.3 & 82.1 \\
    \hline
    \cite{cheng16} & 450D LSTMN with deep attention fusion & 88.5 & 86.3\\
    \cite{parikh16} & 200D decomposable attention model with intra-sentence attention & 90.5 & \bf 86.8\\
    \hline
    \small{\bf Conditional-State-RNN-RNN} & Simple architecture with RNNs without attention & 89.97 & 82.36 \\
    \hline
  \end{tabular}}
  \caption{Comparison with the state-of-the-art in Textual Entailment dataset.}
  \label{table-te}
\end{table*}

\begin{table*}[!ht]
\small
\centering
\resizebox{\textwidth}{!}{\begin{tabular}{|l|r|r|r|r|r|r|r|r|}
    \hline
    \bf Method &\bf P@200 & \bf R@200 & \bf F1@200 & \bf P@50 & \bf R@50 & \bf F1@50 & \bf AVGP & \bf AUC  \\
    \hline
    CDCD \cite{cdcd2014}** & 9.0 & \bf 73.0 & - & 18.0 & 40.0 & - & - & - \\
    BoW \cite{lippi2015} & 8.2 & 51.7 & 14.2 & - & - & - & 0.117 & 0.771\\
    TK \cite{lippi2015} & 9.8 & 58.7 & 16.8 & - & - & - & 0.161 & 0.808\\
    TK+Topic \cite{lippi2015} & \bf 10.5 & 62.9 & \bf 18.0 & - & - & - & 0.178 & 0.823\\
    \hline
    \bf Concat-CNN-CNN  & 9.64 & 61.5 & 15.8 & 17.1 & 27.7 & 19.2 & 0.173 & 0.812 \\
    Conditional-State-Input-RNN-RNN  & 9.56 & 60.0 & 15.6 & 16.6 & 26.9 & 18.5 & 0.162 & 0.801 \\
    \hline
  \end{tabular}}
  \caption{\small Results in Leave-One-Motion-Out mode for Claim Sentence Task. **\newcite{cdcd2014} used a smaller version of the dataset consisting of only 32 motions and also less number of claims. For fair comparison, we also use the same version of dataset as in CDCD and report the results in Appendix ~\ref{sec:supp}. }
  \label{table-lomo-claim}
\end{table*}

\begin{table*}[!ht]
\small
\centering
\resizebox{\textwidth}{!}{\begin{tabular}{|l|r|r|r|r|r|r|r|r|r|r|r|}
    \hline
    \bf Task &\bf P@200 & \bf R@200 & \bf F1@200 & \bf P@50 & \bf R@50 & \bf F1@50 & \bf P@20 & \bf P@10 & \bf P@5 & \bf AVGP & \bf AUC  \\
    \hline
    Claim Sentence Task & 9.64 & 61.5 & 15.8 & 17.1 & 27.7 & 19.2 &  22.4 & 27.9 & 28.5 & 0.173 & 0.812\\
    EXPERT Evidence Task & 9.53 & 64.0 & 14.5& 14.5 & 35.0 & 15.7&  18.6 & 21.1 & 22.5 & 0.160 & 0.750\\
    STUDY Evidence Task & 8.33 & 79.5 & 13.5 & 15.5 & 53.9 & 18.9 &  20.8 & 25.3 & 31.8 & 0.298 & 0.836\\
    \hline
  \end{tabular}}
  \caption{Numbers in Leave-One-Motion-Out mode for all three debating tasks using our approach.}
  \label{table-lomo-all}
\end{table*}

\subsection{Results and Discussion}
Tables ~\ref{table-avgp-summary} and ~\ref{table-auc-summary} report the two top ranking architectures for four datasets based on Test AUC and Test Avg Precision. We find that \textbf{Concat} is the winning architecture variant across majority of the datasets considered. Moreover, the runner-up architecture type \textbf{Conditional-State-Input} is also similar to 'Concat' in the sense that concatenation of context representation is done at the input of the sentence RNN. Now the four datasets we considered are asymmetric in nature as there are significantly fewer contexts (motions or questions) than the targets. Hence the context model does not see enough data for learning and hence, if the learnt context model is fed directly to the hidden state of the target RNN, the improperly learnt context model can play a big role. In contrast if a Concat kind of architecture is used, the linear plus softmax layer can decide on how much importance to give to the context model. Hence, Concat is doing better in this case.

From Textual entailment dataset(which is symmetric in nature), we found that conditional type of architectures are doing better at the Test accuracies. In fact, the winning architecture was \textbf{Conditional-State} with RNN-RNN combo, which did better in terms of test accuracy than the feature based models \cite{snli2015} and one tree-based model \cite{mou15}. However, it came close to the state-of-the-art attention based model \cite{parikh16}. In our work we are empirically evaluating simple architectures for bisequence classification without using more sophisticated tree-based or attention-based models. It is possible that adding attention on top of this will improve the results further.

The \emph{bi-linear} model, is supposed to capture the interaction between the context and target reps via a quadratic form (section ~\ref{sec:biseq-rnn}). For the asymmetric datasets, this is not doing well again due to insufficient data for context. Whereas, it does well for the TE data. However, due to the huge parameter space for bi-linear, training times are considerably higher and requires lower learning rate than other architecture types. The runtimes are comparable for the other architecture variants.

From Table ~\ref{table-lomo-claim}, the main takeaway is that we are the only deep learning based method with zero feature engineering and we have come very close to the state-of-the-art systems \cite{cdcd2014} and \cite{lippi2015}, which are heavily feature-engineered. Here again the winner is a 'Concat' based combination of architecture. Moreover, Tables ~\ref{table-lomo-claim} and ~\ref{table-lomo-all} are the first deep learning zero feature engineered baselines for all argument mining datasets. Appendix ~\ref{sec:supp} contains the details of the exhaustive experiments on all architectures on the different datasets in terms of the best hyperparameters used.



\section{Conclusion}
In this work, we have considered taking up multiple architectures for bisequence classification tasks, for which not much understanding is there in the current literature. In addition to suggesting winning architecture recipes for different kinds of datasets, we have established deep learning based baselines for argument mining tasks with zero feature engineering. As future work, it remains to be seen how adding attention on top of winning simple architectures fare in terms of benchmark performance.

\small
\section{Acknowledgements}
We would like to thank Mitesh M. Khapra for the multiple discussions that we had with him leading to this paper. In addition, we are also grateful to our colleagues at IBM Debating Technologies for the numerous suggestions and feedback at different points of time.

\bibliographystyle{acl}
\bibliography{coling2016}

\appendix
\section{Appendix}
\label{sec:supp}

\begin{table*}[!ht]
\small
\centering
\resizebox{\textwidth}{!}{\begin{tabular}{|l|r|r|r|r|r|r|r|r|}
    \hline
    \bf Method &\bf P@200 & \bf R@200 & \bf F1@200 & \bf P@50 & \bf R@50 & \bf F1@50 & \bf AVGP & \bf AUC  \\
    \hline
    CDCD \cite{cdcd2014} & \bf 9.0 & \bf 73.0 & - & \bf 18.0 & \bf 40.0 & - & - & - \\
    \hline
    \bf Concat-CNN-CNN  & 7.29 & 60.5 & 12.4 & 14.7 & 31.5 & 18.3 & 0.166 & 0.810 \\
    Conditional-State-Input-RNN-RNN  & 6.87 & 58.1 & 11.7 & 13.5 & 29.5 & 17.0 & 0.163 & 0.789 \\
    \hline
  \end{tabular}}
  \caption{\small Results in Leave-One-Motion-Out mode for Claim Sentence Task according to the dataset used by \newcite{cdcd2014}.}
  \label{table-cdcd}
\end{table*}

\begin{table*}[ht]
\footnotesize
\centering
  \resizebox{\textwidth}{!}{\begin{tabular}{|l|l|l|l|r|}
    \hline
    \bf Architecture &\bf Context & \bf Target & \bf Setting & \bf Test AVGP\\
    \hline
    Concat & CBOW & CNN & FilterSize:3,Filters:40 & 0.3 \\
    \hline
    Concat & CBOW & RNN & Cell:GRU,Size:300 & 0.276\\
    \hline
    Concat & RNN & CNN & Cell:LSTM,Size:200,FilterSize:3+4+5,Filters:20 & \bf 0.307\\
    \hline
    Concat & CNN & RNN & Cell:GRU,Size:200,FilterSize:3,Filters:64 & 0.28\\
    \hline
    Concat & RNN & RNN & Cell:GRU,Size:300 & 0.27\\
    \hline
    Concat & CNN & CNN & FilterSize:3+4,Filters:64,L2:0.01 & 0.304\\
    \hline
    Bilinear & CBOW & CNN & FilterSize:3+4+5,Filters:40 & 0.237\\
    \hline
    Bilinear & CBOW & RNN & Cell:GRU,Size:300 & 0.263\\
    \hline
    Bilinear & RNN & CNN & Cell:GRU,Size:200,FilterSize:3+4+5,Filters:64 & 0.254\\
    \hline
    Bilinear & CNN & RNN & Cell:GRU,Size:200,FilterSize:3,Filters:10 & 0.268\\
    \hline
    Bilinear & RNN & RNN & Cell:GRU,Size:200 & 0.263\\
    \hline
    Bilinear & CNN & CNN & FilterSize:3+4,Filters:20 & 0.237\\
    \hline
    Conditional-State & CNN & RNN & Cell:GRU,Size:200,FilterSize:3+4,Filters:100 & 0.248\\
    \hline
    Conditional-State & RNN & RNN & Cell:GRU,Size:200 & 0.266\\
    \hline
    Conditional-Input & CNN & RNN & Cell:GRU,Size:300,FilterSize:3+4+5,Filters:100 & 0.254\\
    \hline
    Conditional-Input & RNN & RNN & Cell:GRU,Size:100 & 0.246\\
    \hline
    Conditional-State-Input & CNN & RNN & Cell:GRU,Size:300,FilterSize:3,Filters:300 & 0.264\\
    \hline
    Conditional-State-Input & RNN & RNN & Cell:GRU,Size:100 & 0.247\\
    \hline
    \multicolumn{3}{|l|}{Concat-Sentence baseline} & Cell:GRU,Size:200 & 0.17\\
    \hline
  \end{tabular}}
  \caption{Best configurations of all architectures on Claim Sentence Dataset tuning based on AVGP}
  \label{table-claim-avgp}
\end{table*}

\begin{table*}[ht]
\footnotesize
\centering
  \resizebox{\textwidth}{!}{\begin{tabular}{|l|l|l|l|r|}
    \hline
    \bf Architecture &\bf Context & \bf Target & \bf Setting & \bf Test AUC\\
    \hline
    Concat & CBOW & CNN & FilterSize:3+4+5,Filters:64,L2:0.01 & 0.863\\
    \hline
    Concat & CBOW & RNN & Cell:GRU,Size:200 & 0.868\\
    \hline
    Concat & RNN & CNN & Cell:GRU,Size:200,FilterSize:3,Filters:40 & 0.867\\
    \hline
    Concat & CNN & RNN & Cell:LSTM,Size:200,FilterSize:3+4+5,Filters:20 & 0.855\\
    \hline
    Concat & RNN & RNN & Cell:GRU,Size:100 & 0.864\\
    \hline
    Concat & CNN & CNN & FilterSize:3,Filters:128,L2:0.01 & \bf 0.873\\
    \hline
    Bilinear & CBOW & CNN & FilterSize:3,Filters:128,L2:0.01,LR:0.0001 & 0.831\\
    \hline
    Bilinear & CBOW & RNN & Cell:GRU,Size:500 & 0.832\\
    \hline
    Bilinear & RNN & CNN & Cell:LSTM,Size:100,FilterSize:3+4,Filters:128,LR:0.00001 & 0.828\\
    \hline
    Bilinear & CNN & RNN & Cell:LSTM,Size:200,FilterSize:3+4+5,Filters:10,LR:0.0001 & 0.857\\
    \hline
    Bilinear & RNN & RNN & Cell:LSTM,Size:300,LR:0.0001 & 0.855\\
    \hline
    Bilinear & CNN & CNN & FilterSize:3+4,Filters:64,L2:0.001,LR:0.0001 & 0.82\\
    \hline
    Conditional-State & CNN & RNN & Cell:GRU,Size:48,FilterSize:3+4+5,Filters:16,LR:0.0001 & 0.873\\
    \hline
    Conditional-State & RNN & RNN & Cell:GRU,Size:100 & 0.86\\
    \hline
    Conditional-Input & CNN & RNN & Cell:GRU,Size:50,FilterSize:3,Filters:50,LR:0.0001 & 0.873\\
    \hline
    Conditional-Input & RNN & RNN & Cell:GRU,Size:200 & 0.86\\
    \hline
    Conditional-State-Input & CNN & RNN & Cell:GRU,Size:300,FilterSize:3+4,Filters:150,LR:0.00001 & 0.856\\
    \hline
    Conditional-State-Input & RNN & RNN & Cell:GRU,Size:200 & 0.862\\
    \hline
    \multicolumn{3}{|l|}{Concat-Sentence baseline} & Cell:GRU,Size:200 & 0.83\\
    \hline
  \end{tabular}}
  \caption{Best configurations of all architectures on Claim Sentence Dataset tuning based on AUC}
  \label{table-claim-auc}
\end{table*}

\begin{table*}[ht]
\small
\centering
  \resizebox{\textwidth}{!}{\begin{tabular}{|l|l|l|l|r|r|}
    \hline
    \bf Architecture &\bf Context & \bf Target & \bf Setting & \bf Test AVGP & \bf Test AUC\\
    \hline
    Concat & CBOW & CNN & FilterSize:3,Filters:40 & 0.239 & 0.81\\
    \hline
    Concat & CBOW & RNN & Cell:GRU,Size:300 & 0.251 & 0.819\\
    \hline
    Concat & RNN & CNN & Cell:LSTM,Size:200,FilterSize:3+4+5,Filters:20 & 0.242 & 0.812\\
    \hline
    Concat & CNN & RNN & Cell:GRU,Size:200,FilterSize:3,Filters:64 & 0.231 & 0.794\\
    \hline
    Concat & RNN & RNN & Cell:GRU,Size:300 & 0.241 & 0.811\\
    \hline
    Concat & CNN & CNN & FilterSize:3+4,Filters:64,L2:0.01 & 0.254 & 0.819\\
    \hline
    Bilinear & CBOW & CNN & FilterSize:3+4+5,Filters:40 & 0.218 & 0.79\\
    \hline
    Bilinear & CBOW & RNN & Cell:GRU,Size:300 & 0.202 & 0.788\\
    \hline
    Bilinear & RNN & CNN & Cell:GRU,Size:200,FilterSize:3+4+5,Filters:64 & 0.219 & 0.789\\
    \hline
    Bilinear & CNN & RNN & Cell:GRU,Size:200,FilterSize:3,Filters:10 & 0.229 & 0.791\\
    \hline
    Bilinear & RNN & RNN & Cell:GRU,Size:200 & 0.214 & 0.788\\
    \hline
    Bilinear & CNN & CNN & FilterSize:3+4,Filters:20 & 0.233 & 0.792\\
    \hline
    Conditional-State & CNN & RNN & Cell:GRU,Size:200,FilterSize:3+4,Filters:100 & 0.226 & 0.797\\
    \hline
    Conditional-State & RNN & RNN & Cell:GRU,Size:200 & 0.254 & \bf 0.832\\
    \hline
    Conditional-Input & CNN & RNN & Cell:GRU,Size:300,FilterSize:3+4+5,Filters:100 & 0.229 & 0.797\\
    \hline
    Conditional-Input & RNN & RNN & Cell:GRU,Size:100 & 0.231 & 0.817\\
    \hline
    Conditional-State-Input & CNN & RNN & Cell:GRU,Size:300,FilterSize:3,Filters:300 & 0.211 & 0.796\\
    \hline
    Conditional-State-Input & RNN & RNN & Cell:GRU,Size:100 & \bf 0.257 & 0.823\\
    \hline
    \multicolumn{3}{|l|}{Concat-Sentence baseline} & Cell:GRU,Size:200 & 0.225 & 0.805\\
    \hline
  \end{tabular}}
  \caption{Performance of all architectures on EXPERT Evidence Dataset}
  \label{table-expert}
\end{table*}

\begin{table*}[ht]
\small
\centering
  \resizebox{\textwidth}{!}{\begin{tabular}{|l|l|l|l|r|r|}
    \hline
    \bf Architecture &\bf Context & \bf Target & \bf Setting & \bf Test AVGP & \bf Test AUC\\
    \hline
    Concat & CBOW & CNN & FilterSize:3,Filters:40 & 0.281 & 0.864\\
    \hline
    Concat & CBOW & RNN & Cell:GRU,Size:300 & 0.279 & 0.851\\
    \hline
    Concat & RNN & CNN & Cell:LSTM,Size:200,FilterSize:3+4+5,Filters:20 & 0.29 & 0.863\\
    \hline
    Concat & CNN & RNN & Cell:GRU,Size:200,FilterSize:3,Filters:64 & 0.262 & 0.829\\
    \hline
    Concat & RNN & RNN & Cell:GRU,Size:300 & 0.28 & 0.842\\
    \hline
    Concat & CNN & CNN & FilterSize:3+4,Filters:64,L2:0.01 & \bf 0.297 & \bf 0.869\\
    \hline
    Bilinear & CBOW & CNN & FilterSize:3+4+5,Filters:40 & 0.271 & 0.831\\
    \hline
    Bilinear & CBOW & RNN & Cell:GRU,Size:300 & 0.202 & 0.788\\
    \hline
    Bilinear & RNN & CNN & Cell:GRU,Size:200,FilterSize:3+4+5,Filters:64 & 0.271 & 0.833\\
    \hline
    Bilinear & CNN & RNN & Cell:GRU,Size:200,FilterSize:3,Filters:10 & 0.254 & 0.839\\
    \hline
    Bilinear & RNN & RNN & Cell:GRU,Size:200 & 0.257 & 0.84\\
    \hline
    Bilinear & CNN & CNN & FilterSize:3+4,Filters:20 & 0.275 & 0.835\\
    \hline
    Conditional-State & CNN & RNN & Cell:GRU,Size:200,FilterSize:3+4,Filters:100 & 0.254 & 0.835\\
    \hline
    Conditional-State & RNN & RNN & Cell:GRU,Size:200 & 0.267 & 0.861\\
    \hline
    Conditional-Input & CNN & RNN & Cell:GRU,Size:300,FilterSize:3+4+5,Filters:100 & 0.245 & 0.838\\
    \hline
    Conditional-Input & RNN & RNN & Cell:GRU,Size:100 & 0.28 & 0.854\\
    \hline
    Conditional-State-Input & CNN & RNN & Cell:GRU,Size:300,FilterSize:3,Filters:300 & 0.257 & 0.839\\
    \hline
    Conditional-State-Input & RNN & RNN & Cell:GRU,Size:100 & 0.25 & 0.849\\
    \hline
    \multicolumn{3}{|l|}{Concat-Sentence baseline} & Cell:GRU,Size:200 & 0.236 & 0.844\\
    \hline
  \end{tabular}}
  \caption{Performance of all architectures on STUDY Evidence Dataset}
  \label{table-study}
\end{table*}

\begin{table*}[ht]
\small
\centering
  \resizebox{\textwidth}{!}{\begin{tabular}{|l|l|l|l|r|r|}
    \hline
    \bf Architecture &\bf Context & \bf Target & \bf Setting & \bf Test AVGP & \bf Test AUC\\
    \hline
    Concat & CBOW & CNN & FilterSize:3,Filters:40 & 0.162 & 0.735\\
    \hline
    Concat & CBOW & RNN & Cell:GRU,Size:300 & \bf 0.187 & 0.74\\
    \hline
    Concat & RNN & CNN & Cell:LSTM,Size:200,FilterSize:3+4+5,Filters:20 & 0.15 & 0.727\\
    \hline
    Concat & CNN & RNN & Cell:GRU,Size:200,FilterSize:3,Filters:64 & 0.119 & 0.66\\
    \hline
    Concat & RNN & RNN & Cell:GRU,Size:300 & 0.171 & 0.705\\
    \hline
    Concat & CNN & CNN & FilterSize:3+4,Filters:64,L2:0.01 & 0.179 & \bf 0.74\\
    \hline
    Bilinear & CBOW & CNN & FilterSize:3+4+5,Filters:40 & 0.129 & 0.672\\
    \hline
    Bilinear & CBOW & RNN & Cell:GRU,Size:300 & 0.119 & 0.656\\
    \hline
    Bilinear & RNN & CNN & Cell:GRU,Size:200,FilterSize:3+4+5,Filters:64 & 0.122 & 0.676\\
    \hline
    Bilinear & CNN & RNN & Cell:GRU,Size:200,FilterSize:3,Filters:10 & 0.131 & 0.681\\
    \hline
    Bilinear & RNN & RNN & Cell:GRU,Size:200 & 0.149 & 0.688\\
    \hline
    Bilinear & CNN & CNN & FilterSize:3+4,Filters:20 & 0.129 & 0.712\\
    \hline
    Conditional-State & CNN & RNN & Cell:GRU,Size:200,FilterSize:3+4,Filters:100 & 0.122 & 0.681\\
    \hline
    Conditional-State & RNN & RNN & Cell:GRU,Size:200 & 0.171 & 0.739\\
    \hline
    Conditional-Input & CNN & RNN & Cell:GRU,Size:300,FilterSize:3+4+5,Filters:100 & 0.141 & 0.713\\
    \hline
    Conditional-Input & RNN & RNN & Cell:GRU,Size:100 & 0.184 & 0.729\\
    \hline
    Conditional-State-Input & CNN & RNN & Cell:GRU,Size:300,FilterSize:3,Filters:300 & 0.186 & 0.726\\
    \hline
    Conditional-State-Input & RNN & RNN & Cell:GRU,Size:100 & 0.169 & 0.714\\
    \hline
  \end{tabular}}
  \caption{Performance of all architectures on WikiQA Dataset}
  \label{table-wikiqa}
\end{table*}

\makeatletter
\setlength{\@fptop}{5pt}
\makeatother

\begin{table*}[ht]
\centering
  \resizebox{\textwidth}{!}{\begin{tabular}{|l|l|l|l|r|r|r|}
    \hline
    \bf Architecture &\bf Context & \bf Target & \bf Setting & \bf TrainAcc(\%) & \bf ValidAcc(\%) & \bf TestAcc(\%)\\
    \hline
    Concat & CBOW & CNN & FilterSize:3+4+5,Filters:128 & 74.33 & 69.43 & 68.44\\
    \hline
    Concat & CBOW & RNN & Cell:LSTM,Size:400 & 72.75 & 69.4 & 69.02\\
    \hline
    Concat & RNN & CNN & Cell:GRU,Size:200,FilterSize:3+4+5,Filters:20 & 74.01 & 69.34 & 68.96\\
    \hline
    Concat & CNN & RNN & Cell:GRU,Size:200,FilterSize:3,Filters:20 & 72.93 & 69.99 & 69.69\\
    \hline
    Concat & RNN & RNN & Cell:LSTM,Size:200 & 72.74 & 69.96 & 69.46\\
    \hline
    Concat & CNN & CNN & FilterSize:3+4+5,Filters:64 & 74.55 & 69.49 & 68.97\\
    \hline
    Bilinear & CBOW & CNN & FilterSize:3+4,Filters:128 & 83.86 & 77.1 & 77.07\\
    \hline
    Bilinear & CBOW & RNN & Cell:GRU,Size:300 & 84.78 & 79.07 & 78.19\\
    \hline
    Bilinear & RNN & CNN & Cell:GRU,Size:500,FilterSize:2+3+4+5,Filters:200 & 84.42 & 77.68 & 77.18\\
    \hline
    Bilinear & CNN & RNN & Cell:GRU,Size:500,FilterSize:2+3+4+5,Filters:200 & 83.71 & 78.72 & 78.6\\
    \hline
    Bilinear & RNN & RNN & Cell:GRU,Size:1000,LR:0.0001 & 84.91 & 81.1 & 80.3\\
    \hline
    Bilinear & CNN & CNN & FilterSize:3+4,Filters:128,LR:0.0001 & 84.51 & 76.58 & 76.81\\
    \hline
    Conditional-State & CNN & RNN & Cell:GRU,Size:500,FilterSize:3,Filters:500 & 87.77 & 80.87 & 80.81\\
    \hline
    Conditional-State & RNN & RNN & Cell:GRU,Size:500 & \bf 89.97 & \bf 82.38 & \bf 82.36\\
    \hline
    Conditional-Input & CNN & RNN & Cell:GRU,Size:500,FilterSize:3+4,Filters:250 & 87.36 & 80.81 & 81.1\\
    \hline
    Conditional-Input & RNN & RNN & Cell:GRU,Size:500 & 89.02 & 81.45 & 80.92\\
    \hline
    Conditional-State-Input & CNN & RNN & Cell:GRU,Size:500,FilterSize:3,Filters:500 & 85.78 & 80.05 & 79.61\\
    \hline
    Conditional-State-Input & RNN & RNN & Cell:GRU,Size:500 & 89.03 & 81.93 & 81.38\\
    \hline
  \end{tabular}}
  \caption{Performance of all architectures on Textual Entailment Dataset}
  \label{table-te}
\end{table*}

\end{document}